\documentclass{article}
\PassOptionsToPackage{numbers}{natbib}
\usepackage[dblblindworkshop,final]{neurips_2026}

\usepackage[utf8]{inputenc}
\usepackage[T1]{fontenc}
\usepackage{hyperref}
\usepackage{url}
\usepackage{booktabs}
\usepackage{amsfonts}
\usepackage{amsmath}
\usepackage{nicefrac}
\usepackage{microtype}
\usepackage{xcolor}
\usepackage{wrapfig}

\newif\ifcamerareadychanges
\camerareadychangesfalse
\DeclareRobustCommand{\crchange}[1]{{\ifcamerareadychanges\color{blue}\fi#1}}
\title{What Transfers Across Protein Pockets?\\Experience Reuse in Molecular
Optimization for Agentic Structure-Based Drug Design}
\workshoptitle{AI for Drug Discovery: Bridging the Translation Gap (NeurIPS 2026)}

\author{%
  \crchange{Zaifei Yang$^{1}$}\thanks{\crchange{Equal contribution.}}\quad
  \crchange{Weiyu Chen$^{2}$}\footnotemark[1]\quad
  \crchange{Yaqing Wang$^{3}$}\thanks{\crchange{Corresponding author.}}\quad
  \crchange{James Kwok$^{1}$} \\
  \crchange{$^1$The Hong Kong University of Science and Technology} \\
  \crchange{$^2$Hong Kong Institute of AI for Science, City University of Hong Kong} \\
  \crchange{$^3$Beijing Institute of Mathematical Sciences and Applications} \\
  \crchange{\texttt{zyangea@connect.ust.hk}, \texttt{weiyu.chen@cityu.edu.hk}} \\
  \crchange{\texttt{wangyaqing@bimsa.cn}, \texttt{jamesk@cse.ust.hk}}
}

\begin{document}
\maketitle

\begin{abstract}
LLM agents optimize a ligand in a protein pocket by editing it step by step and docking each edit.
Recent agents first spend extra dockings probing the complex with controlled edits; guidance induced from these probes is reported to help more than the chemical knowledge the model already carries.
The probes are then discarded with the instance that produced them, so the obvious next step is to keep them in an experience base, letting a new pocket inherit what earlier ones already paid for.
Such a base is read through a \emph{key}: the record properties under which past edits are pooled and a new one is looked up. 
Using the full log of one such run over the CrossDocked2020 test-set pockets, we ask whether its keys recur on a pocket the base has never seen and still separate improving from worsening edits.
\crchange{Using} the contacts at the site being edited, the action, and the round as \crchange{a} key, the base separates them sharply.
Within our analysis, this separation is largely explained by how much the edit changed the molecule's size, the action, and the round, none of which describes the target.
That is not because the target is irrelevant: naming it outright separates them several times more sharply.
For a pocket absent from the base, transfer must rely on a description rather than its identity.
Among the exact-match descriptions we evaluate, the more specific ones rarely recur, whereas the coarser ones provide little separation; the signals that do transfer primarily describe the edit and the molecule it produced rather than the target.
Under the representations and lookup scheme studied here, an experience base may be more useful for prioritizing an agent's own proposals for docking than for directly prescribing what to build.
\end{abstract}

\section{Introduction}
\label{sec:intro}
LLM agents that iteratively refine a ligand inside a protein pocket are a promising new approach for ligand optimization in structure-based drug design (SBDD) \citep{liddia,cidd,probe}.
The most recent of these, PROBE, adds a probing stage that measures how a particular pocket-ligand complex responds to controlled edits, and reports that guidance induced from those measured responses is better than prior chemical knowledge inside the agent\citep{probe}.
But \crchange{measuring edit responses in the pocket requires} extra dockings on top of those the optimization itself needs, and the per-instance edit manual it produces is never carried beyond that instance.
In contrast, agents outside chemistry write what a run learns into an experience base and retrieve it on later tasks \citep{reflexion, voyager,awm,expreuse,skillpro}.
Thus, a promising direction is to accumulate the edit responses, so that a new optimization could \crchange{start} from evidence earlier runs already bought.

That direction depends on how reliably measurements from one pocket remain informative on another.
We test it on \crchange{an} experience base formed by a complete run log of one full PROBE run over the CrossDocked2020 test-set pockets (Section~\ref{sec:setup}).
Such a base is read through a \emph{key}: a tuple of record properties under which past records are pooled and a candidate edit is looked up. 
A useful key should have three things hold at once: (i) it should utilize information from both the pocket and the ligand, (ii) it recurs on a pocket the base has never seen, and (iii) it still separates what an edit does there.

The records already supply a key that appears to meet all three: the contacts at the edited atom, the action, and the round.
It recurs on almost every edit and separates improving from worsening ones sharply, but \crchange{for the two \textsf{Replace} actions, the gap becomes indistinguishable from zero after matching the action, heavy-atom change, and round}.
As the only description of pockets in this key, the contact features add little detectable separation after the edit properties are held equal.
Naming the pocket separates outcomes several times more sharply than that key does,
but a pocket the base has never seen must be represented through transferable descriptors rather than identity. Among the discretized, exact-match descriptions we evaluate, those that separate outcomes recur on few held-out pockets, whereas those that recur broadly add little separation.
Thus, none of the evaluated keys meets all three requirements at once. The keys that transfer most consistently describe properties of the edit or its product more than properties of the pocket.
\crchange{We discuss candidate prioritization as a possible use of this base.}

\section{Related work}
\label{sec:related}
\paragraph{LLM agents for molecular optimization in structure-based drug design.}
Recent pipelines differ in what they ask the LLM to do, but \crchange{all keep the measured experience within the current instance}.
CIDD \citep{cidd} has several LLM roles that propose edited molecules directly. 
LIDDIA \citep{liddia} uses the LLM as a planner and hands the edit to an algorithmic executor.
PROBE \citep{probe} spends extra dockings per instance on the LLM proposed controlled edits, measuring how the pocket responds, and turning those responses into per-instance optimization guidance. 
In all three cases, \crchange{information learned about the pocket remains within the current optimization instance rather than being reused on a new pocket}.

\paragraph{Experience reuse in LLM agents.}
Outside chemistry, agents already keep what a run learns and read it back later, as a verbal account of what went wrong \citep{reflexion}, as a library of reusable skills \citep{voyager}, or as workflows distilled from past trajectories \citep{awm}.
\crchange{These systems also face questions of transfer. In SBDD, we examine a specific form of transfer: whether edit outcomes measured on one protein pocket remain informative on another.}

\section{Setup}
\label{sec:setup}
An agent for molecular optimization in SBDD is given a protein pocket and a ligand, and edits the ligand step-by-step so that the molecule binds the pocket more tightly and has better druggability.
Following \citep{probe}, we score a molecule $m$ by affinity $A(m) = -\mathrm{Vina}(m)$ \citep{vina} and calculate $N_{m}$ as the number of heavy atoms. 
We define $\Delta A = A(m')-A(m)$, where
$m'$ is the optimized molecule after one optimization step, and $\Delta N=N_{m'}-N_{m}$ is the change in heavy-atom count.

\paragraph{The experience base.}
Our experience base comes from a single full run of the PROBE workflow (Appendix~\ref{app:probe}) on the  CrossDocked2020 test-set pockets \citep{crossdocked}.
For each pocket, PROBE optimized 10 initial ligand molecules generated by MolCRAFT \citep{molcraft}, proceeding through a probe phase (labeled round 0) followed by five rounds of design-make-test-analyze (labeled rounds 1--5).
We recorded each \crchange{successfully} built-and-docked edit by the agent as a record in the experience base, ultimately generating 14,341 records, including 5,398 from the probe phase and 8,943 from DMTA edits.
Each record in the experience base includes the following properties: (i) pocket, (ii) ligand, (iii) round, (iv) editing action, (v) four categorical constraints on the fragment to introduce, (vi) pocket-ligand contact features, and (vii) the outcome $\Delta A, \Delta N$. Appendix~\ref{app:records} details each property.

\paragraph{Reading the base on a held-out pocket.}
We use the experience base to predict whether a candidate edit will improve affinity. 
A key $\kappa$, a tuple of record properties, selects the subset of records that agree with it on those properties. Writing $n^{+}_\kappa$ and $n^{-}_\kappa$ for the number of records in that subset with $\Delta A > 0$ and with $\Delta A < 0$, the base predicts \textsf{improve} when $n^{+}_\kappa \ge n^{-}_\kappa$ and \textsf{worsen} otherwise. 
Every \crchange{held-out evaluation} below splits at the pocket, not at the record: for each pocket in turn, the base is built from the records of the other pockets and used to label the held-out pocket's records, so every record is labelled by a base that never saw its pocket.

\section{Transferable keys primarily describe the edit}
\label{sec:exp}

Anything the base carries has to be carried by a key, so we test whether the evaluated keys meet all three requirements at once.
The key the records already supply recurs and separates, but its separation is largely associated with properties of the edit (Section~\ref{sec:q1}).
The pocket does change what an edit gains, but the pocket descriptions studied here show a trade-off between recurrence and separation (Section~\ref{sec:q2}).

\subsection{\crchange{The initial separation largely reflects edit properties}}
\label{sec:q1}
The records already supply a key, $\kappa_0 = (\text{pocket-ligand contact features},\ \text{editing action},\ \text{round})$.
Read through $\kappa_0$, the base separates improving from worsening edits sharply: the records it labels \textsf{improve} improve far more often than the ones it labels \textsf{worsen} (Table~\ref{tab:match}, row~1).
Of its three fields, only the contact features describe the pocket, so to the extent that the gap is explained by the action, the round, or the heavy-atom change, the prediction does not require pocket information.

\crchange{Stratifying by an edit property compares records within groups matched on that property.}
We average the gap over groups of records that agree on that property, by the Cochran--Mantel--Haenszel stratified risk difference~\citep{cochran,mh},
\begin{equation}
\label{eq:matched}
\widehat{\Delta} = \sum_{c} w_c \bigl(\hat p^{\,\mathrm{imp}}_{c} - \hat p^{\,\mathrm{wor}}_{c}\bigr),
\qquad
w_c \propto \frac{n^{\mathrm{imp}}_{c}\, n^{\mathrm{wor}}_{c}}
                 {n^{\mathrm{imp}}_{c} + n^{\mathrm{wor}}_{c}},
\end{equation}
where $\hat p^{\,\mathrm{imp}}_{c}$ and $n^{\mathrm{imp}}_{c}$ are the improvement rate and count of
the records predicted \textsf{improve} in group $c$, and $\hat p^{\,\mathrm{wor}}_{c}$ and
$n^{\mathrm{wor}}_{c}$ are the same for \textsf{worsen}.
A large $\widehat{\Delta}$ indicates separation beyond the held-equal property; a small one indicates that the prediction adds little separation beyond that property.


\begin{wraptable}{r}{0.46\textwidth}
  \vspace{-\intextsep}
  \caption{Gap in improvement rate between records labelled \textsf{improve} and
  \textsf{worsen}, averaged over groups that agree on the listed properties
  (Eq.~\eqref{eq:matched}); $\Delta N$ is held at its exact value. Intervals are
  $95\%$ percentile intervals over $2{,}000$ bootstrap resamples of the pockets.
  Rows are comparable within a panel, not across.}
  \label{tab:match}
  \centering\footnotesize
  \setlength{\tabcolsep}{4pt}
  \begin{tabular}{@{}rlrr@{}}
    \toprule
    \# & Held equal & $\widehat{\Delta}$ (pp) & 95\% CI\\
    \midrule
    & \multicolumn{3}{@{}l}{\emph{all four actions}} \\
    1 & nothing                   & $+53.3$ & $[+47.6,+57.9]$ \\
    2 & $\Delta N$                & $+2.7$  & $[-3.0,+8.3]$   \\
    \midrule
    & \multicolumn{3}{@{}l}{\emph{the two \textsf{Replace} actions only}} \\
    3 & action                    & $+36.1$ & $[+28.9,+42.6]$ \\
    4 & action, $\Delta N$        & $+10.8$ & $[+4.3,+17.5]$  \\
    5 & action, $\Delta N$, round & $+0.3$  & $[-5.1,+6.3]$   \\
    \bottomrule
  \end{tabular}
  \vspace{-\intextsep}
\end{wraptable}

Much of the gap can be explained by the heavy-atom change.
The action largely fixes that count, and the Vina score tracks it~\citep{hopkins2004}, so we hold $\Delta N$ equal first, though it is not a field of $\kappa_0$.
Doing so \crchange{reduces the gap to an estimate whose interval includes zero} over all four actions (row~2).
However, that comparison cannot separate the heavy-atom change from the action name.
The base labels every \textsf{Grow} record \textsf{improve} and every \textsf{Delete\_Group} record \textsf{worsen}, and those two actions move $\Delta N$ in opposite directions.

Holding the action equal drops those two: their groups have records on one side only, so Eq~\eqref{eq:matched} gives them no weight.
Rows~3--5 therefore rest on the two \textsf{Replace} actions alone.
The action by itself still leaves a large gap (row~3), adding $\Delta N$ leaves a small one (row~4), and adding the round closes it (row~5).
The contact features are then the only field of $\kappa_0$ left, and what they leave has an interval covering zero (row~5).
For the two \textsf{Replace} actions, the observed separation is largely accounted for by the heavy-atom change, the action, and the round; for the other two, it is closely aligned with the action. None of these properties describes the pocket.

\subsection{The evaluated pocket keys trade recurrence for separation}
\label{sec:q2}
If pockets do not change what an edit gains, a cross-pocket base has nothing pocket-specific to hold.

\begin{table}[t]
  \caption{\crchange{Sixteen keys, grouped by information source. \emph{Recurs} is the fraction of held-out records with an exact match in the base. $\omega^2$ measures in-sample separation of residual outcomes (Eq.~\eqref{eq:resid}), corrected for group count. Predictions use the majority residual sign among matching records. \emph{Beats $\Delta N$} is the fraction of evaluated pockets with higher accuracy than a $\Delta N$ lookup, with a $95\%$ binomial interval; \emph{Lift} is the mean accuracy difference. Identity keys (rows~1--3) cannot recur. $\kappa_0$ is bold; HBD/HBA denote hydrogen-bond donor/acceptor counts. Appendix~\ref{app:pocketdesc} gives pocket-descriptor definitions, group counts, and evaluated pockets.}}
  \label{tab:keys}
  \centering\small
  \begin{tabular}{rlrrrrr}
    \toprule
    \# & Key & Recurs & $\omega^2$ & Beats $\Delta N$ & 95\% CI & Lift (pp) \\
    \midrule
    & \multicolumn{6}{l}{\emph{context}} \\
     1 & input ligand                & 0\%   & 27.6\% & --- & --- & --- \\
     2 & pocket, action       & 0\%   & 20.6\% & --- & --- & --- \\
     3 & pocket               & 0\%   & 12.0\% & --- & --- & --- \\
    \midrule
    & \multicolumn{6}{l}{\emph{pocket description}} \\
     4 & pocket descriptors, all six & 17\%  & 11.8\% & 35\% & $[13,58]$ & $-5.5$ \\
     5 & \ \ burial                  & 100\% & 0.6\%  & 39\% & $[29,49]$ & $-3.4$ \\
     6 & \ \ hydrophobic fraction    & 100\% & 0.2\%  & 42\% & $[32,52]$ & $-1.0$ \\
    \midrule
    & \multicolumn{6}{l}{\emph{edit}} \\
     7 & \textbf{contact features, action, round} & 94\% & 2.5\% & 52\% & $[41,62]$ & $+1.2$ \\
     8 & contact features, action    & 98\%  & 1.6\%  & 57\% & $[47,67]$ & $+2.8$ \\
     9 & action                      & 100\% & 0.9\%  & 49\% & $[39,60]$ & $+0.1$ \\
    10 & contact features            & 99\%  & 0.2\%  & 52\% & $[41,62]$ & $-0.1$ \\
    \midrule
    & \multicolumn{6}{l}{\emph{edit specification}} \\
    11 & fragment constraints        & 90\%  & 6.9\%  & 63\% & $[54,73]$ & $+5.1$ \\
    \midrule
    & \multicolumn{6}{l}{\emph{ligand}} \\
    12 & $\Delta$rotatable bonds     & 100\% & 5.8\%  & 85\% & $[78,92]$ & $+8.0$ \\
    13 & $\Delta$aromatic atoms      & 99\%  & 4.0\%  & 70\% & $[61,79]$ & $+6.8$ \\
    14 & $\Delta$rings               & 100\% & 3.1\%  & 91\% & $[86,97]$ & $+8.1$ \\
    15 & $\Delta$HBD                 & 100\% & 1.2\%  & 72\% & $[63,81]$ & $+2.8$ \\
    16 & $\Delta$HBA                 & 100\% & 0.8\%  & 53\% & $[43,63]$ & $+0.1$ \\
    \bottomrule
  \end{tabular}
  \vskip -0.2in
\end{table}

\textbf{Every key is scored on what the heavy-atom change and the round leave unexplained.}
A key carrying the heavy-atom change on top of the action and the round would cut the records too finely to recur on a new pocket.
So we fit the heavy-atom change and the round out of the outcome instead, and let the action stand as a key of its own:
\begin{equation}
\label{eq:resid}
\varepsilon = \Delta A - \bigl(\beta_0 + \beta_1 \Delta N + \beta_2 \Delta N^{2}
+ \beta_3\,\mathrm{round} + \beta_4 A(m)\bigr)
\end{equation}
fitted by ordinary least squares and refitted in every held-out fold; 
$A(m)$ is the parent's own affinity.
The fit takes out the trend in each field, not everything the field says; a key looks up exact values, so $\Delta N$ and the round can still separate what is left.
From here the outcome is $\varepsilon$, not $\Delta A$.

\textbf{A key is scored twice: on how sharply it separates, and on whether it still does so on a pocket the base has never seen.}
A key $\kappa$ partitions the $n$ records into $k$ groups. Writing $n_g$ and $\bar\varepsilon_g$
for the size and mean of group $g$ and $\bar\varepsilon$ for the overall mean, let
$B$ \crchange{be} the between-group and $W$ \crchange{be} the within-group sums of squares and $s^2$ \crchange{be} the within-group variance.
We define $\omega^2$ as the percentage of the variance of $\varepsilon$ the partition explains, in
its \crchange{bias-corrected} form \citep{olejnik2003},
\[
B = \sum_g n_g\bigl(\bar\varepsilon_g - \bar\varepsilon\bigr)^2, \qquad
W = \sum_g \sum_{i \in g}\bigl(\varepsilon_i - \bar\varepsilon_g\bigr)^2, \qquad
s^2 = \frac{W}{n-k}, \qquad
\omega^2 = \frac{B-(k-1)\,s^2}{B+W+s^2}
\]
The $(k-1)s^2$ correction is needed because $B$ grows with $k$ even for a key that carries no information. $\omega^2$ is a point estimate and carries no interval and is an in-sample quantity, computed over the whole corpus and not validated out of fold.
The second score is out of fold, against a lookup on $\Delta N$ alone: Section~\ref{sec:q1} found that the heavy-atom change explains much of the initial separation, so a useful key should improve on what it already predicts.

\textbf{The five categories differ in what a key may use: the pocket, the ligand, or the edit.}
(i) \emph{Context} keys name the pocket or the input ligand outright.
(ii) \emph{Pocket description} keys describe the pocket structurally: residue count, four residue-class fractions, and burial.
(iii) \emph{Edit} keys are $\kappa_0$ and its parts; the contact features are the only pocket information they carry.
(iv) The \emph{edit specification} key is the four fragment constraints, all the agent says about the fragment to introduce.
(v) \emph{Ligand} keys describe what the edit did to the ligand.

\textbf{How much an edit gains depends on the pocket.}
Pocket identity separates $\varepsilon$ several times more sharply than $\kappa_0$ does (Table~\ref{tab:keys}, rows~3 and~7), so there is pocket-dependent structure for a cross-pocket base to hold.
Refining that partition by the action, or by the ligand the optimization started from, separates it more sharply again (rows~1 and~2).
The base can describe a pocket it has never seen but cannot name one, so rows~1--3 \crchange{provide in-sample reference partitions, not upper bounds on unseen-pocket performance}.

\textbf{The evaluated pocket descriptions do not combine broad recurrence with strong separation.}
All six descriptors together \crchange{approach pocket identity's in-sample separation} (row~4).
But they form almost one group per pocket (Table~\ref{tab:pocketdesc}), so few held-out records find a group the base already holds, and the key can be scored on few pockets at all.
Coarsening the descriptors reverses the trade-off.
Burial, hydrophobic fraction, and the contact features recur everywhere but explain little outcome variation (rows~5, 6, 10).
Additionally, rows~4, 5, 6, and 10 do not outperform the baseline in this evaluation.

\textbf{The keys that retain separation on a new pocket primarily describe the ligand rather than the pocket.}
The fragment constraints the agent writes, and four properties of the ligand the edit produced, recur almost everywhere and beat the baseline on most pockets (rows~11--15).
Four of them separate outcomes more sharply than $\kappa_0$ does (rows~11--14).
$\Delta$HBA recurs, but \crchange{its mean accuracy lift is only $0.1$ pp, and the interval for its fraction of pockets beating the baseline includes $50\%$} (row~16).
None of these keys describes the pocket. In this setting, the transferable signal therefore comes mainly from properties that can be computed for the current edit, rather than from retrieved target-specific experience.

\section{Discussion and limitations}
\label{sec:disc}

\textbf{The evaluated keys reveal a recurrence--specificity trade-off.}
The pocket does carry structure (Section~\ref{sec:q2}), but the exact-match representations studied here capture it only through relatively fine partitions, which limits recurrence.
Of the three things a useful key needs, incorporating pocket information creates the main difficulty in this analysis: keys with little or no pocket information meet the other two requirements more comfortably (Table~\ref{tab:keys}, rows~11--15).

\textbf{The results favor proposal ranking over direct edit prescription.}
The keys that retain separation on a new pocket contain little pocket information, and inside one pocket they may not need it for ranking.
The base could rank those candidates by how its own records with the same key value turned out, allowing the agent to prioritize which ones to dock.
Every key value it reads is fixed before that docking is paid for: the fragment constraints when the agent writes the edit, the ligand keys once the candidate is built (Appendix~\ref{app:records}).
Directly prescribing what to build would require a transferable representation that conditions effectively on a previously unseen pocket; we do not identify such a key among the representations evaluated in Section~\ref{sec:q2}.
\crchange{Candidate ranking, closed-loop optimization gains, and docking savings have not been evaluated here.}

\textbf{Limitations.} 
(i) The evidence comes from one agent pipeline (PROBE) run once per instance, one generator of initial ligands (MolCRAFT), one scoring function (AutoDock Vina), the CrossDocked2020 test pockets, one set of six structural pocket descriptors, and one family of keys: categorical, exact-match lookups over discretised record properties.
A different description of the pocket, or a key that pools by similarity rather than exact match, is untested; our conclusions therefore apply to the representations and lookup family evaluated here rather than to cross-pocket experience reuse in general.
\crchange{In particular, combining pocket descriptors with the strongest ligand-change keys remains untested.}
(ii) The corpus is assembly- and policy-selected: only edits that built and docked become records, and every edit was proposed under the agent's own site map and EditManual, which already rules out the edits it judges harmful.
Those are also the edits such an agent would be looking up at run time, so the corpus matches the setting the base is meant for; an agent that proposes edits without a pocket-fitted manual is untested.
\crchange{Differences in retained record counts, action distributions, and chemical diversity across pockets are not separately characterized.}
\crchange{(iii) The endpoint is a Vina score, not experimental binding. Intervals are unadjusted for multiple key comparisons, so marginal differences are exploratory.}

\clearpage
\begin{ack}
This research was supported in part by the Research Grants Council of the Hong Kong Special Administrative Region (Grants 16202523 and 16213126). Yaqing Wang is sponsored by Beijing Nova Program.

\paragraph{Competing interests.} The authors declare no competing interests.
\end{ack}

\appendix
\bibliographystyle{plainnat}
\bibliography{ref}

\newpage

\section{The PROBE workflow}
\label{app:probe}

PROBE \citep{probe} optimizes one pocket-ligand complex at a time.
The input ligand comes from a 3D generative model and is docked into the pocket.
PROBE scores a molecule on two objectives at once: affinity, from its Vina score, and
druggability, a score $D(m)$ built from QED and synthetic accessibility.
Its premise is that the agent should measure how this particular pocket answers a few controlled
edits before it commits to an optimization direction.

\paragraph{Site map.}
PROBE decomposes the input ligand with BRICS \citep{brics}.
For each fragment it collects three streams of signal: which protein contacts the fragment makes,
from PLIP \citep{plip}; how the fragment fits the pocket geometrically; and property signals
covering ligand efficiency and known problematic groups.
A diagnosis LLM reads these, writes a short profile of the complex, and picks three or four
fragments as editable \emph{sites}.
It labels each site \textsc{synergy} (one edit can plausibly improve both objectives),
\textsc{tension} (improving one is likely to hurt the other), or \textsc{liability} (a structural
liability to repair whatever the trade-off).
Those labelled sites are the \emph{site map}.
It is built once, from the input complex, and never changes.

\paragraph{Probing.}
A planner LLM proposes three strategies --- affinity-first, druggability-first, co-optimization
--- each naming target sites and an edit prescription.
A generator LLM turns each strategy into four structured edits: three that apply it at high,
medium and low magnitude, and one counterfactual that reverses it as a control.
That is twelve probe edits per input ligand.
\crchange{All twelve are submitted for assembly, docking, and scoring; only successfully built-and-docked edits enter our corpus.}
This probe stage is the whole of what we label round $0$.

\paragraph{EditManual.}
An analyzer summarizes the \crchange{available probe} outcomes.
For each strategy it reports whether the high--medium--low response is monotone improvement, an
activity cliff, saturation, or flat and negative, and whether reversing the strategy makes the
scores worse, better or unchanged; a strategy is credible only if the forward probes improve and
the counterfactual does not.
A manual-construction LLM then writes the \emph{EditManual}.
For each site it records a pharmacophore role, the actions allowed and forbidden there together
with the size, polarity, flexibility and shape any later fragment must satisfy, and a blacklist of
structural features.
It also records which sites may be edited independently and which edit directions are mutually
exclusive.

\paragraph{Multi-agent optimization.}
Five DMTA rounds follow, rounds $1$--$5$ in our labelling.
In each round an affinity agent and a druggability agent each draft one localized edit under the
site map, the EditManual and the edit history.
The affinity agent prefers \textsc{synergy} sites, the druggability agent prefers
\textsc{liability} sites, and both first resolve their chosen site to its current atom indices,
because the molecule changes across rounds while the site map does not.
The two drafts are cross-reviewed against the manual, and a co-optimization agent reconciles them
into a third.
All three are built, docked and labelled against the intent that produced them.
The edit history is extended, and any EditManual entry an outcome contradicts is tightened.

\paragraph{Selection.}
At the end of round $0$ and of every DMTA round, PROBE promotes one candidate to be the next
parent.
Candidates are ranked by their joint gain over the input ligand $m_0$,
\begin{equation}
\label{eq:vol}
S_{\mathrm{bal}}(m) =
\begin{cases}
\dfrac{A(m)-A(m_0)}{|A(m_0)|}\cdot\dfrac{D(m)-D(m_0)}{D(m_0)}
  & \text{if } A(m) > A(m_0) \text{ and } D(m) > D(m_0),\\[2pt]
0 & \text{otherwise,}
\end{cases}
\end{equation}
so a candidate that fails to improve both objectives \crchange{receives a score of zero}.
After the last round PROBE returns the maximizer of $S_{\mathrm{bal}}$ over a pool holding the
input ligand, the \crchange{successfully evaluated} probes, and every molecule the optimization produced.

\paragraph{Fragment-assembly engine.}
One engine builds every edit, in the probe stage and in the rounds alike.
A structured edit names an edit atom, an action, and four categorical constraints on the fragment
to introduce (Appendix~\ref{app:records}).
\textsf{Delete\_Group} removes the fragment rooted at the edit atom and caps the open valence.
The other three actions need a fragment, which the engine draws from a library of $8{,}505$ BRICS
fragments \citep{brics} of ChEMBL Phase-I compounds.
Each library fragment carries a precomputed QED, synthetic accessibility, chiral-centre count,
rotatable-bond count, molecular weight and discrete shape category, so the constraints become a
database query.
The engine restricts the library to the matching subset, ranks it by
$0.5\,\mathrm{QED}(f) + 0.5\,\mathrm{SA}(f) - 0.15\,N_{\text{chiral}}(f)$ with $N_{\text{chiral}}$
the number of chiral centres, and uses RDKit to attach the top-ranked fragment and
sanitize the product.

\section{Corpus record schema}
\label{app:records}

A record is one structured edit that the fragment-assembly engine built and that Vina scored.
The record's \emph{parent} $m$ is the molecule the edit was applied to: at round $0$ the input
ligand, at rounds $1$--$5$ the candidate promoted at the end of the previous round
(Appendix~\ref{app:probe}).
Its \emph{product} $m'$ is the molecule the engine returned.
Fields (i)--(vi) are fixed by the time the agent writes the edit, before the product exists.
Of the outcomes in (vii), only $\Delta A$ needs the docking; the ligand counts can be read as soon
as the engine returns the product.

\paragraph{(i)--(iii) Identifiers.}
The \textsf{pocket} is one of the $93$ CrossDocked2020 test-set pockets \citep{crossdocked}, each
from a different protein, carried as its CrossDocked identifier.
The \textsf{ligand} says which of that pocket's ten MolCRAFT input ligands \citep{molcraft} the
optimization started from.
The pair (\textsf{pocket}, \textsf{ligand}) is an \emph{instance}; there are $930$.
The \textsf{round} is the round the edit was made in (Section~\ref{sec:setup}).
The six pocket descriptors (Appendix~\ref{app:pocketdesc}) are properties of the pocket, not
fields of the record; they join on \textsf{pocket}.

\paragraph{(iv) Editing action.}
One of four actions.
\textsf{Delete\_Group} removes the fragment rooted at the edit atom and caps the open valence.
\textsf{Grow} attaches a library fragment at the edit atom.
\textsf{Replace\_Terminal\_Group} and \textsf{Replace\_Sidechain\_or\_Ring} swap a library fragment
for the terminal group, or for the sidechain or ring system, rooted there.
We record the edit atom index but never key on it, because an index means nothing in another
molecule.

\paragraph{(v) Fragment constraints.}
The three actions that add a fragment carry four constraints:

\begin{center}\small
\begin{tabular}{ll}
\toprule
field & values \\
\midrule
\textsf{size}        & \textsf{Small} (MW $<150$) / \textsf{Medium} ($150$--$250$) / \textsf{Large} ($>250$) / \textsf{Any} \\
\textsf{polarity}    & \textsf{Hydrophilic} / \textsf{Neutral} / \textsf{Lipophilic} / \textsf{Any} \\
\textsf{flexibility} & \textsf{Rigid} ($0$--$1$ rotatable bonds) / \textsf{Flexible} ($>1$) / \textsf{Any} \\
\textsf{shape}       & \textsf{Disc-like} / \textsf{Rod-like} / \textsf{Sphere-like} / \textsf{Any} \\
\bottomrule
\end{tabular}
\end{center}

\noindent A \textsf{Delete\_Group} record introduces no fragment and carries no constraint values,
so row~11 of Table~\ref{tab:keys} is scored on the other three actions.
The engine has a fifth field, formal charge, that the agent never set; we do not record it.
Within a constraint cell the engine returns a fixed fragment (Appendix~\ref{app:probe}), so which
molecule gets built is decided by these four fields and the edit atom, not by the agent's choice of
fragment.

\paragraph{(vi) Pocket-ligand contact features.}
PLIP \citep{plip} annotates every ligand heavy atom of the docked \emph{parent} complex with the
protein contacts it forms.
We join the contact types at the edit atom, in a fixed order, into one categorical label, and write
\textsf{NONE} where the atom forms no detected contact.
The label drops residue identity, the number of contacts of each type, and all distances and
angles.
\textsf{NONE} covers $61\%$ of the records; \textsf{hbond} and \textsf{hydrophobic} follow, and no
other value reaches $5\%$.

\paragraph{(vii) Outcome.}
Vina scores of parent and product give $A(m)$ and $\Delta A$; heavy-atom counts give $\Delta N$
(Section~\ref{sec:setup}).
From the same two molecules we take the changes in rotatable-bond count, aromatic-atom count, ring
count, and hydrogen-bond donor and acceptor counts; these five are the \emph{ligand} keys of
Table~\ref{tab:keys}.
We compute all five with RDKit at its default definitions, and a key reads each at
its exact integer value.

\section{Pocket descriptors}
\label{app:pocketdesc}

Six descriptors describe a pocket.
Each is computed from the receptor structure and the reference ligand shipped with the
CrossDocked2020 test split \citep{crossdocked}, and from nothing the run produces, so a key can
read them on a pocket the base has never seen.
The residue set is every residue with a heavy atom within $10$~\AA{} of a reference-ligand heavy
atom.
Five descriptors summarize it: its residue count, and its fraction of residues in each of four
classes.

\begin{center}\small
\begin{tabular}{ll}
\toprule
class & residues \\
\midrule
hydrophobic & \textsc{ala}, \textsc{val}, \textsc{leu}, \textsc{ile}, \textsc{phe}, \textsc{met}, \textsc{trp}, \textsc{pro}, \textsc{cys} \\
charged     & \textsc{asp}, \textsc{glu}, \textsc{lys}, \textsc{arg}, \textsc{his} \\
aromatic    & \textsc{phe}, \textsc{tyr}, \textsc{trp}, \textsc{his} \\
polar       & \textsc{ser}, \textsc{thr}, \textsc{asn}, \textsc{gln}, \textsc{tyr}, \textsc{cys}, \textsc{his} \\
\bottomrule
\end{tabular}
\end{center}

\noindent The classes overlap and omit \textsc{gly}, so the four fractions do not sum to one.
Burial, the sixth descriptor, counts receptor heavy atoms within $8$~\AA{} of a reference-ligand
heavy atom, per reference-ligand heavy atom.
We cut each descriptor into tertiles over the $93$ pockets.
No descriptor is near-constant: residue count runs from $24$ to $106$, burial from $3.5$ to
$25.8$, and every class fraction varies at least twofold.

\begin{table}[h]
  \caption{Keys built from the pocket, alone and joined to the edit, with the edit keys and pocket
  identity for reference. One row is one key, measured as in Table~\ref{tab:keys}. \emph{Cells} is
  the number of groups the key forms, and \emph{Pockets} the number of held-out pockets on which
  \emph{Beats $\Delta N$} can be estimated at all; it falls below $93$ only for the two keys built
  from all six descriptors, where few held-out records find a group the base already holds. Rows
  marked $\dagger$ also appear in Table~\ref{tab:keys}.}
  \label{tab:pocketdesc}
  \centering\small
  \begin{tabular}{lrrrrr}
    \toprule
    Key & $\omega^2$ & Cells & Beats $\Delta N$ & 95\% CI & Pockets \\
    \midrule
    \multicolumn{6}{l}{\emph{the edit}} \\
    action$^{\dagger}$                              & 0.9\%  & 4   & 49\% & $[39,60]$ & 93 \\
    contact features$^{\dagger}$                    & 0.2\%  & 14  & 52\% & $[41,62]$ & 93 \\
    contact features, action$^{\dagger}$            & 1.6\%  & 46  & 57\% & $[47,67]$ & 93 \\
    \midrule
    \multicolumn{6}{l}{\emph{the pocket alone}} \\
    residue count                                   & 0.1\%  & 3   & 40\% & $[30,50]$ & 93 \\
    hydrophobic fraction$^{\dagger}$                & 0.2\%  & 3   & 42\% & $[32,52]$ & 93 \\
    charged fraction                                & 0.3\%  & 3   & 51\% & $[40,61]$ & 93 \\
    aromatic fraction                               & 0.3\%  & 3   & 29\% & $[20,38]$ & 93 \\
    polar fraction                                  & 0.1\%  & 3   & 48\% & $[38,59]$ & 93 \\
    burial$^{\dagger}$                              & 0.6\%  & 3   & 39\% & $[29,49]$ & 93 \\
    pocket descriptors, all six$^{\dagger}$         & 11.8\% & 84  & 35\% & $[13,58]$ & 17 \\
    \midrule
    \multicolumn{6}{l}{\emph{the pocket and the action}} \\
    residue count, action                           & 1.2\%  & 12  & 56\% & $[46,66]$ & 93 \\
    hydrophobic fraction, action                    & 1.9\%  & 12  & 49\% & $[39,60]$ & 93 \\
    charged fraction, action                        & 1.7\%  & 12  & 44\% & $[34,54]$ & 93 \\
    aromatic fraction, action                       & 1.4\%  & 12  & 53\% & $[43,63]$ & 93 \\
    polar fraction, action                          & 1.2\%  & 12  & 43\% & $[33,53]$ & 93 \\
    burial, action                                  & 1.6\%  & 12  & 58\% & $[48,68]$ & 93 \\
    pocket descriptors, all six, action             & 19.5\% & 332 & 29\% & $[8,51]$  & 17 \\
    \midrule
    \multicolumn{6}{l}{\emph{the pocket and the edit}} \\
    residue count, contact features, action         & 2.0\%  & 108 & 56\% & $[46,66]$ & 93 \\
    hydrophobic fraction, contact features, action  & 2.9\%  & 108 & 54\% & $[44,64]$ & 93 \\
    charged fraction, contact features, action      & 2.5\%  & 110 & 56\% & $[46,66]$ & 93 \\
    aromatic fraction, contact features, action     & 2.3\%  & 111 & 52\% & $[41,62]$ & 93 \\
    polar fraction, contact features, action        & 2.5\%  & 108 & 49\% & $[39,60]$ & 93 \\
    burial, contact features, action                & 2.8\%  & 110 & 55\% & $[45,65]$ & 93 \\
    \midrule
    \multicolumn{6}{l}{\emph{pocket identity}} \\
    pocket$^{\dagger}$                              & 12.0\% & 93  & ---  & ---       & --- \\
    \bottomrule
  \end{tabular}
\end{table}

The $(k-1)s^2$ correction in $\omega^2$ (Section~\ref{sec:q2}) is what keeps a key from looking
informative just because it forms many groups.
Permuting $\varepsilon$ across records and scoring every key again leaves $\omega^2$ at $0.1\%$ or
below, the two keys built from all six descriptors included.
\crchange{This check supports the group-count correction, but does not establish held-out predictive value.}

\end{document}